%% file: root.tex
\documentclass[letterpaper, 10 pt, conference]{ieeeconf}  

\IEEEoverridecommandlockouts                              

\usepackage{graphics} 
\usepackage{epsfig} 
\usepackage{mathptmx} 
\usepackage{times} 
\usepackage{amsmath} 
\usepackage{amssymb}  
\usepackage[dvipsnames]{xcolor}

\usepackage{booktabs}
\usepackage{tikz}
\usetikzlibrary{patterns}
\usepackage{pgfplots}
\pgfplotsset{compat=1.7}
\usepackage{pgfplotstable}
\usepackage{latexsym}
\usepackage{microtype}
\usepackage[utf8]{inputenc}
\usepackage[T1]{fontenc}
\usepackage{url}
\usepackage{siunitx}
\input{addons}

\title{\LARGE \bf
Contextual Emotion Recognition using Large Vision Language Models
}

\author{Yasaman Etesam and \"{O}zge Nilay Yal\c{c}{\i}n and Chuxuan Zhang and Angelica Lim \\
  Simon Fraser University, BC, Canada \\
  \texttt{yetesam@sfu.ca,  oyalcin@sfu.ca, cza152@sfu.ca, angelica@sfu.ca } \\
}

\begin{document}

\maketitle
\thispagestyle{empty}
\pagestyle{empty}

\input{latex/Sections/abstract}

\section{Introduction}

\input{latex/Sections/introduction}
\input{latex/Sections/contributions}

\section{Related Work}
\input{latex/Sections/related-work}

\section{Methodology}
\input{latex/Sections/method}

\section{Experiments}

\input{latex/Sections/experiments}

\section{Limitations}
\input{latex/Sections/limitations}

\section{Conclusion}
\input{latex/Sections/conclusion}



\bibliographystyle{IEEEtran} 
\bibliography{custom}
\addtolength{\textheight}{-12cm}   

\end{document}

%% file: addons.tex
\usepackage{graphicx}
\usepackage{wrapfig}
\usepackage{multirow}

\usepackage[dvipsnames]{xcolor}

\newcommand{\mypara}[1]{\vspace{2pt}\noindent\textbf{#1}}

%% file: latex/Sections/abstract.tex
\begin{abstract}
``How does the person in the bounding box feel?" Achieving human-level recognition of the apparent emotion of a person in real world situations remains an unsolved task in computer vision. Facial expressions are not enough: body pose, contextual knowledge, and commonsense reasoning all contribute to how humans perform this emotional theory of mind task. In this paper, we examine two major approaches enabled by recent large vision language models: 1) image captioning followed by a language-only LLM, and 2) vision language models, under zero-shot and fine-tuned setups. We evaluate the methods on the Emotions in Context (EMOTIC) dataset and demonstrate that a vision language model, fine-tuned even on a small dataset, can significantly outperform traditional baselines. The results of this work aim to help robots and agents perform emotionally sensitive decision-making and interaction in the future.



\end{abstract}


%% file: latex/Sections/introduction.tex


Our ability to recognize emotions allows us to understand one another, build successful long-term social relationships, and interact in socially appropriate ways. Equipping virtual agents and robots with emotion recognition capabilities can help us improve and facilitate human-machine interactions \cite{picard2000affective}. However, emotion recognition systems today still suffer from poor performance \cite{barrett2019emotional} due to the complexity of the task. This innate and seemingly effortless capability requires understanding of the causal relations, contextual information, social relationships as well as theory of mind, which are unresolved problems in affective computing research. Many image-based emotion recognition systems focus solely on using facial or body features \cite{pantic2000expert,schindler2008recognizing}, which can lead to a low accuracy in the absence of contextual information \cite{barrett2011context,barrett2017emotions}. 

In the past few years, the affective computing research community has been moving towards creating datasets and building models that include or make use of contextual information. The EMOTIC dataset, for instance, incorporates contextual and environmental factors for apparent emotion recognition in still images~\cite{kosti2019context}. The inclusion of contextual information beyond facial features is found to significantly improve the accuracy of the emotion recognition models \cite{le2022global,mittal2020emoticon}. However, using this information to infer the emotions of others requires commonsense knowledge and high-level cognitive capabilities such as reasoning and theory of mind which are missing from traditional emotion recognition models~\cite{ong2019computational}. 

Another limitation of traditional emotion recognition models is that many of them are trained and tested on the same dataset~\cite{lopez2016dependence}. This stands in contrast to the challenge of generalization, where robots may perform poorly in novel situations~\cite{jang2022bc}. In this study, we employ zero-shot models and observe their performance in unseen scenarios. Additionally, we demonstrate how results can be enhanced through fine-tuning. Both LLMs and LVMs are evaluated for this purpose.

Large language models (LLMs) that are based on the transformer architecture \cite{vaswani2017attention} have been shown to excel at natural language processing (NLP) tasks \cite{brown2020language, chowdhery2022palm}, offering a way to achieve emotional theory of mind through linguistic descriptors. LLMs gained success in increasing accuracy and efficiency in NLP problems including multimodal tasks such as visual question answering \cite{antol2015vqa} and caption generation \cite{vinyals2015show}. Recently, they have been also used in commonsense reasoning \cite{sap2019socialiqa,bisk2020piqa,li2022systematic}, emotional inference \cite{mao2022biases} and theory of mind \cite{sap2022neural} tasks, however their capabilities on emotional theory of mind in visual emotion recognition tasks have not been explored. 
\input{latex/tables/teaser}

Vision language models (VLMs) integrate natural language processing with visual comprehension to generate text from visual inputs and are capable of performing a variety of visual recognition tasks. VLMs learn intricate vision-language correlations from large-scale image-text pair datasets, enabling zero-shot predictions across a range of visual recognition tasks~\cite{zhang2024vision}. Despite their success in tasks like image classification~\cite{pratt2023does} and object detection~\cite{long2023fine}, their capability in contextual emotion recognition has not yet been explored.

In this paper, we focus on a multi-label, contextual emotional theory of mind task by utilizing the embedded knowledge in large language models (LLMs) and vision language models (VLMs). To the best of our knowledge, this is the first evaluation of VLMs in the contextual emotion recognition task.


%% file: latex/tables/teaser.tex
\begin{figure}[t]
\centering
\includegraphics[width=1\linewidth]{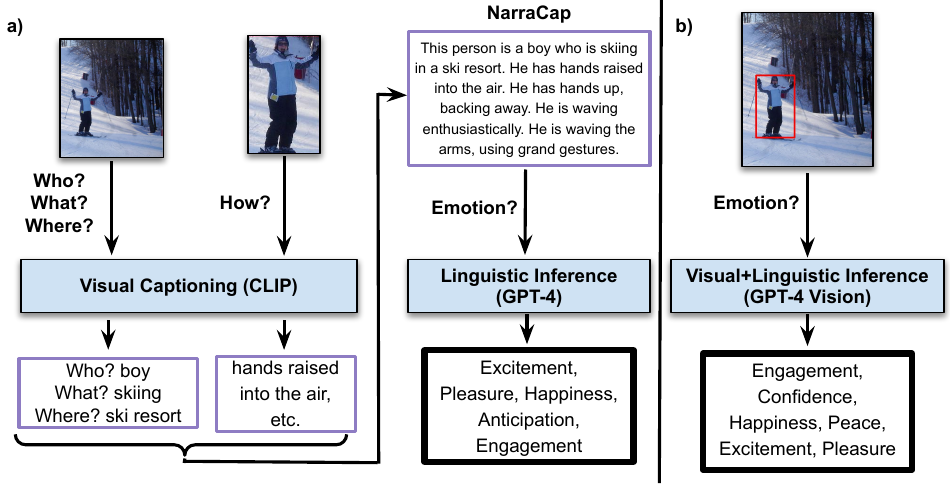}
\vspace{-7mm}
\caption{
In this paper, we evaluated two distinct zero-shot approaches:
a) using CLIP to generate captions from images, followed by providing these captions alongside a prompt to large language models (LLMs) to obtain emotion labels.
b) directly providing the image and a prompt to vision language models (VLMs) and requesting the emotion labels.}

\label{fig:teaser}
\vspace{-4mm}
\end{figure}

%% file: latex/Sections/contributions.tex
The contributions of this paper are as follows:



\begin{itemize}
    \item Presenting a new state-of-the-art open source method for affect estimation in the wild, using VLM fine-tuning
    
    \item Proposing zero-shot approaches for contextual emotion recognition to explore generalizability for robotics 

    \item Evaluating the effectiveness of a) captioning + LLM, versus b) VLM approaches for emotion recognition


\end{itemize}

%% file: latex/Sections/related-work.tex
\emph{Emotional theory of mind in context.} Work in emotional theory of mind (in this paper, also referred to as emotion recognition) has been focusing on the inclusion of contextual information in addition to the facial or posture information in recent years. Early datasets such as HAPPEI \cite{dhall2013finding} proposed, for instance, emotion annotation for groups of people. More recently, the EMOTIC dataset \cite{kosti2019context} was developed as multilabel dataset containing 26 categories of emotions, 18,316 images and 23,788 annotated people. The related emotion recognition task is to provide a list of emotion labels that matches those chosen by annotators, responding to the question of, ``How does the person in the bounding box feel?". In approximately 25\% of the person targets, the face was not visible, underscoring the role of context in estimating the emotion of a person in the image. The phrase "emotional theory of mind" is used here to clarify that we are not estimating the sentiment or emotional content of an image, but estimating the emotion of a particular person contained in the image. Note that we do not claim to perform felt emotion recognition, but apparent emotion recognition as perceived by labelers.

\emph{Vision-based approaches for contextual emotion estimation}. A number of computer vision approaches have been developed in response to the release of the EMOTIC dataset. The EMOTIC baseline~\cite{kosti2019context} uses a CNN to extracts features from the target human, as well as the entire image.
Subsequent fusion methods incorporated body and context visual information \cite{huang2021emotion} at global or local scales \cite{le2022global}, investigated contextual videos \cite{lee2019context}, or worked to improve subsets of the EMOTIC dataset, such as \cite{thuseethan2021boosting} which considered the photos only including two people. In PERI \cite{mittel2023peri}, attention is modulated at various levels within their feature extraction network. In~\cite{li2021human}, relational region-level analysis was employed, while~\cite{hoang2021context} utilized visual relationship between the main target and adjacent objects.
To the extent of our knowledge, the current best approach was Emoticon proposed by \cite{mittal2020emoticon}, which explored the use of graph convolutional networks and image depth map. This approach can further improved by adding CCIM~\cite{yang2023context} plug-in.
Overall, the latest results leave room for improvement, and no pretrained models are available for roboticists, with the exception of EMOTIC.

\emph{Large language models and theory of mind.} Recent investigations into large language models (LLMs) have uncovered some latent capabilities for social intelligence, including some sub-tasks on emotion inference \cite{mao2022biases}. Emotional theory of mind tasks using language tend to focus on appraisal based reasoning on emotions, inferring based on a sequence of events. For instance, among other social intelligence tasks, \cite{sap2022neural} explored how a language model could respond to an emotional event, e.g. ``Although Taylor was older and stronger, they lost to Alex in the wrestling match. How does Alex feel?" Their findings suggested some level of social and emotional intelligence in LLMs.


\emph{Natural language and emotional theory of mind.} Language is a fundamental element in emotion and it plays a crucial role in emotion perception~\cite{lindquist2015role, lieberman2007putting, lindquist2013s}. 
In this work, in order to apply LLMs, we use the body of work in English literature that discusses how writers use empathy to describe characters, their actions and external cues, and ``seek to evoke feelings in readers employing the powers of narrativity." \cite{keen2015narrative}. Writing textbooks such as The Emotion Thesaurus \cite{puglisi2019emotion} provide sets of visual cues or actions that support emotional evocation, as a guide for readers to imagine the most relevant features of the person in the scene. For example, to evoke curiosity, a writer may narrate, "she tilted the head to the side, leaning forward, eyebrows furrowing" or "she raised her eyebrows and her body posture perked up." 


\emph{Generalization and zero-shot learning.}
A major challenge in the field of robotics is generalization, enabling robots to adapt to new and unforeseen scenarios~\cite{jang2022bc}. Effective training on a labeled dataset often results in satisfactory performance only when evaluations are conducted on similar datasets. This is because the emphasis on minimizing training error tends to make machines capture all the correlations present in the training data, rather than understanding the actual causation~\cite{lopez2016dependence}. In the real world, the distribution of objects across different categories can exhibit a long-tailed pattern, where some categories are represented by a large number of training images, while others have few or no images at all~\cite{wang2019survey, liu2018generalized, pourpanah2022review}. This disparity makes it challenging to identify all possible correlations. A crucial goal in visual emotion recognition is to improve the ability to recognize instances of previously unseen visual classes, an approach known as zero-shot learning~\cite{abderrahmane2018haptic, socher2013zero}. In our work, we leverage the embedded knowledge within large language and vision language models to address the visual emotional theory of mind task, aiming to perform this task without training on a specific dataset.

%% file: latex/Sections/method.tex

In this study, we compare two general approaches: a) 2-phased image captioning and large language models, with b) end-to-end vision language models.

\subsection{Image Captioning and Large Language Model}


In this first method, we use a two-phased approach to first generate a caption of the image, then use an LLM for linguistic reasoning to perform emotion inference (see Fig. \ref{fig:teaser}). Our captioning method is called Narrative Captioning (NarraCap), and we compare it to a state-of-the-art ExpansionNet~\cite{hu2022expansionnet} captioning. 


\subsubsection{Narrative Captioning}
Our zero-shot Narrative Captioning (NarraCap) makes use of templates and the vision language model CLIP~\cite{radford2021learning}. First, given an image with the bounding box of a person, we extract the cropped bounding box and pass it along with a gender/age category (baby girl, baby boy, girl, boy, man, woman, elderly man, elderly woman) to CLIP to understand \textbf{who} is in the picture. Next, we pass the entire image to understand \textbf{what} is happening in the image by selecting the action with the maximum probability. The action list comprises 848 different actions extracted from the Kinetics-700 \cite{smaira2020short}, UCF-101 \cite{soomro2012ucf101}, and HMDB datasets \cite{kuehne2011hmdb}.   

We then add the \textbf{how} aspect of the image by passing the cropped bounding box through CLIP, along with 889 signals (available on our website\footnote{https://yasaman-etesam.github.io/Contextual-Emotion-Recognition/}) filtered from over 1000 social signals derived from a guide on writing about emotion \cite{puglisi2019emotion}. 
Using trial and error, we found that the best approach comes from selecting signals that, when paired with an image in CLIP, returns a probability higher than $mean+9*std$ of the class label scores.
To provide additional context, we use 224 environmental descriptors from a writer's guide to urban \cite{puglisi2016urban} and rural \cite{puglisi2016rural} settings to describe \textbf{where} the person in the scene is located.
The prompts we selected for CLIP are as follows: `\textit{A photo of a(n) [gender/age/location]}', and `\textit{A photo of a person who is(has) [action/physical signals]}'. 
Examples of narrative captions (NarraCap) can be found in Fig.~\ref{tab:qualitative}. 

\subsubsection{ExpansionNet Captioning}
We also evaluate a baseline captioning method, ExpansionNet \cite{hu2022expansionnet}, a fast end-to-end trained model for image captioning. 
The model achieves state of the art performance over the MS-COCO 2014 captioning challenge, and was used as a backbone for a recent approach trained on EMOTIC~\cite{de_Lima_Costa_2023_CVPR}, and serves as a  baseline to our NarraCap approach.
\subsubsection{Caption to Emotion using LLM (Zero Shot)}
Following captioning, we provide the caption, along with a prompt, to GPT-4.
The prompt 
asks for the top emotion labels understood from the caption:
"\textit{<caption> From suffering, pain, }
\textit{[...], and sympathy, pick the top labels that this person is feeling at the same time.}"
We also utilized an open-source LLM, Mistral 7B~\cite{jiang2023mistral}, which incorporates grouped-query attention~\cite{ainslie2023gqa} and sliding window attention~\cite{child2019generating, beltagy2020longformer} techniques to address common LLM limitations such as computational power and memory requirements. Mistral, with its 7 billion parameters, outperforms the best released 34-billion-parameter model, Llama 1~\cite{touvron2023llama}, in reasoning tasks. The combination of a relatively low parameter count and high performance makes Mistral a potential option for applications in robotics. The prompt for Mistral is as follows: "\textit{<caption> From suffering, pain, }
\textit{[...], and sympathy, the top labels that this person is feeling at the same time are:}"
\subsubsection{Mistral (Fine-Tuned)}Fine-tuning LLMs~\cite{min2021metaicl, ouyang2022training, liu2022few} has been demonstrated as an effective strategy to improve their performance. In this paper, using quantization, LoRA~\cite{hu2021lora, dettmers2024qlora} and NarraCap, we finetune Mistral on the emotion recognition task. 
For Mistral, we experiment by fine-tuning on the Emotic validation set and augmentation.

\subsection{End-to-End Vision Language Models}

We next explore 3 vision language models (VLMs):  CLIP, a closed-source (GPT-4) and an open source (LLaVA) VLM. We also study the effect of prompt engineering, as well as fine-tuning on the open-source model (LLaVA).

\subsubsection{CLIP (Zero Shot)}
CLIP jointly trains an image encoder and a text encoder by maximizing the cosine similarity between related (image, text) pairs and minimizing the cosine similarity between the irrelevant pairs:
\begin{equation}
\begin{aligned}
\text{logits} &= \text{np.dot}(I_e, T_e^T) * \text{np.exp}(t) \\
\end{aligned}
\end{equation}
where $I_e$ is image feature embeddings and $T_e$ is the text feature embeddings. 
CLIP can be used to perform zero shot classification by comparing distances between an image and various texts in a multimodal embedding space. We used the images from EMOTIC and compared the distances with each of the emotion labels and selected the six (average number of ground truth labels in validation set) labels with highest probabilities as our labels. 

\subsubsection{GPT-4 Vision and LLaVA (Zero Shot)}
GPT-4 Vision is a proprietary model from OpenAI that can provide text-based responses given an image and text input. Large Language and Vision Assistant (LLaVA)~\cite{liu2023visual} is an open source multi-purpose multimodal model designed by combining CLIP's visual encoder~\cite{radford2021learning} and LLAMA's language decoder~\cite{touvron2023llama}. 
The model is fine-tuned end-to-end on the language-image instruction-following data generated using GPT-4~\cite{openai2023gpt4}.

\subsubsection{LLaVA (Fine-Tuned)}
 We use the EMOTIC data to finetune LLaVA with LoRA~\cite{hu2021lora}\footnote{https://github.com/haotian-liu/LLaVA} on the emotion recognition task.
 We experiment by fine-tuning LLaVA on the EMOTIC training set ($17077$ images, and $23706$ individuals), EMOTIC validation set ($2087$ images, and $3330$ individuals), and on a small dataset, created by selecting $100$ images at random from the validation set.
Furthermore, we perform data augmentation by shuffling the ground truth labels for each image in the validation set and using each image $3$ times with different shuffled labels.

\subsubsection{Prompt Engineering}

We used the images from EMOTIC and a text prompt: "\textit{From suffering, pain, [...], and sympathy, pick the top labels that the person in the red bounding box is feeling at the same time.}" It has been shown that prompt engineering (e.g. chain of thought~\cite{wei2022chain}) is an effective way to improve results. We also tested a prompt which included definitions of the emotions and specified a number of labels to output.

%% file: latex/Sections/experiments.tex
Our experiment is focused on the EMOTIC dataset, which covers 26 different labels. The related emotion recognition task is to provide a list of emotion labels that matches those chosen by annotators.
The training set (70\%) was annotated by 1 annotator, where validation (10\%) and test (20\%) sets were annotated by 5 and 3 annotators respectively. 
While previous work on emotion recognition tasks~\cite{kosti2019context, mittal2020emoticon, mittel2023peri}, utilize the mean Average Precision (mAP) as a metric, in this work, outputs are textual descriptions indicating the labels the person is feeling, rather than probabilities. Therefore, we could not employ mAP, instead, we used precision, recall, F1 score, hamming loss, which demonstrates the average rate at which incorrect labels are predicted for a sample, and subset accuracy, which requires the predicted set of labels for a sample to exactly match the actual set of labels. These metrics are implemented using the scikit-learn library \cite{scikit-learn}. 
We compare the following methods: 

\mypara{EMOTIC} Along with the dataset,~\cite{kosti2019context} introduced a two-branched CNN-based network baseline. 
The first branch extracts body related features and the second branch extracts scene-context features. Then a fusion network combines these features and estimates the output. For the EMOTIC dataset, using the provided code\footnote{https://github.com/Tandon-A/emotic} and thresholds calculated from the validation set, we obtained the output labels on the test set and then calculated the target metrics. This approach is the only traditional method with reproducible code.\\ 
\mypara{Emoticon} Motivated by Frege's principle \cite{resnik1967context}, \cite{mittal2020emoticon} proposed an approach by combining three different interpretations of context. They used pose and face features (context1), background information (context2), and interactions/social dynamics (context3). They used a depth map to model the social interactions in the images. Later, they concatenate these different features and pass it to fusion model to generate outputs~\cite{mittal2020emoticon}.
Unfortunately, the code for this project was not made available by the authors, and we could not reproduce the reported results. Consequently, we cannot provide a reliable comparison with this approach.

\mypara{Random} We consider selecting either $6$ (average number of labels per person in validation set) emotions randomly from all possible labels (\textit{Rand}) or selecting $6$ labels randomly where the weights are determined by the number of times each emotion is repeated in the validation set (\textit{Rand(W)}). 

\mypara{Majority} This Majority baseline selects the top $6$ most common emotions in the validation set (engagement, anticipation, happiness, pleasure, excitement, confidence) as the predicted labels for all test images (\textit{Maj}). 

\mypara{CLIP} For the CLIP model, we employed the \textit{clip-vit-base-patch32}. While utilizing the \textit{clip-vit-large-patch14-336} model did enhance the F1 score for the clip-only method to $19.60$, its use significantly increased processing time, particularly for generating NarraCap captions. Therefore, to maintain consistency in our reporting and efficiency in our processing, we present results using the \textit{clip-vit-base-patch32} model. The prompt used here is: \textit{"The person in the red bounding box is feeling \{emotion label\}"}. 
Additionally, we employed Grad-CAM~\cite{selvaraju2017grad}, to generate saliency maps, allowing us to visually highlight the areas within images that significantly influenced the model's decisions.

\mypara{Captions+GPT-4} After generating captions, we pass the captions to \textit{gpt-4 (gpt-4-0613)}. We utilized GPT-4 with the temperature parameter set to 0 and the maximum token count set to 256. Additionally, the frequency penalty, presence penalty, and top\_p were configured to 0, 0, and 1, respectively. While adjusting these parameters could potentially enhance the model's performance, we refrained from hyperparameter tuning for this task due to associated costs.

\mypara{GPT-4 Vision} Using \textit{gpt-4-vision-preview}, we input EMOTIC test images, with parameters set similarly to GPT-4. In this experiment, we tested both the prompt mentioned in Section 3.2.4, and also the inclusion of label definitions provided by EMOTIC to GPT, and requesting the six most likely labels.

\mypara{LLaVA} As GPT4-Vision, we tested both the prompt mentioned in Section 3.2.4, and also the inclusion of label definitions provided by EMOTIC to GPT, and requesting six most likely labels. LLAVA fine-tuning was performed on four A40 48GB GPUs.

\mypara{Mistral} We used huggingface\footnote{https://huggingface.co} to run and finetune Mistral on a RTX 3090 Ti GPU. We used maximum new tokens of $256$ and repetition penalty equal to $1.15$.


\section{Results and Discussion}

\input{latex/tables/table1_pdf}

\input{latex/tables/ap_per_category}

\input{latex/tables/clip-map}

The results for zero shot methods are shown in Table \ref{tab:res}, and example images with captions in Fig.~\ref{tab:qualitative}.

We observe that fine-tuning LLaVA with an augmented validation set provides the best overall F1 score. In Fig.~\ref{tab:qualitative}, we can see that LLaVA fine-tuned on the validation set predicts more labels than when trained on the training set. 
One explanation is that the number of annotators for the training and test sets is $1$ and $3$, respectively. Since we utilize the combined labels predicted by all annotators, the average number of ground truth labels per person is higher in the test set ($4.42$) than in the training set ($1.96$). This discrepancy leads the model trained on the training set to predict fewer labels than what is present in the test ground truth, causing the model to predict cautiously with high precision but miss many labels, resulting in low recall. To address this issue, we attempted fine-tuning on the validation set, which has 5 annotators and an average of $6.157$ ground truth labels per person.
We also experimented with fine-tuning on a small dataset, selecting $100$ images at random from the validation set. This was to demonstrate that using a minimal amount of data can still yield reasonable results with vision language models (VLMs).
Future work could try to balance the average number of labels in the training and test set. We also see that simple augmentation of the validation set, by shuffling the labels, improves performance. This may be due to the model learning that label ordering is not an important factor in the text output.

In Fig.~\ref{tab:qualitative}, we observe that the EMOTIC baseline tends to predict many more labels than the other methods, which reduces its precision and overall F1 score. For an application where choosing a precise emotion label is more important than predicting all possible labels, fine-tuned LLaVA on the training set may be the most useful model.

It is evident that CLIP, which underperforms as indicated in Table~\ref{tab:res}, misinterprets certain images, such as mistakenly attributing the emotion of embarrassment to a woman at the beach (Fig. \ref{tab:qualitative}). A deeper analysis using Grad-Cam-generated saliency maps (as seen in Fig. \ref{tab:clip-map}.1) offers a plausible explanation: CLIP may inaccurately associate images displaying bare skin with embarrassment. Additionally, CLIP seems to exhibit spurious correlations of body language, predicting emotions like surprise and fear in response to raised arms (Fig.\ref{tab:clip-map}.2), or sadness from the positioning of hands near the face (Fig.~\ref{tab:clip-map}.3).

In the captioning combined with GPT-4 analysis, NarraCap proves to be more effective than ExpNet in aiding GPT-4's understanding of emotions. However, it ranks as the second-best zero-shot approach. GPT-4 Vision with prompt engineering emerges as the top performer among zero-shot methods, surpassing EMOTIC, which was trained on the EMOTIC training set.  

\mypara{How does captioning + LLM compare to the end-to-end VLM approach?}
In addition to our experiments in Table \ref{tab:res}, we performed an additional study on a smaller test set. Yang et al. \cite{yang2023contextual} recruited an annotator fluent in North American English to manually generate captions for 387 images, encompassing 14 negative emotion categories: suffering, annoyance, pain, sadness, anger, aversion, disquietment, doubt/confusion, embarrassment, fear, disapproval, fatigue, disconnection, and sensitivity. This focus on negative emotions stemmed from their comparatively poor recognition across all methods tested, relative to positive emotions. The outcomes for all methodologies applied to this dataset are detailed in Table~\ref{tab:overview_f1_scores}. On this smaller, challenging test set, our best proposed zero-shot captioning + LLM approach (NarraCap+GPT-4) resulted in an F1 score of $\textbf{26.19}$. The GPT4-Vision zero-shot VLM approach attained an F1 score of $\textbf{35.79}$. This disparity appears large; however, leveraging human-generated captions with GPT-4 (LLM) achieved an F1 score of $\textbf{34.17}$. This indicates that while the \textit{automatic captioning + LLM} method does not reach the VLM performance, human-level captioning when coupled with LLMs provides nearly comparable performance and outperforms the traditional EMOTIC baseline.

\input{latex/tables/human_set}

\mypara{How do different prompts affect the results?} Selecting an appropriate prompt for LLMs and VLMs is crucial for optimizing their performance. However, the same prompt can affect different models in varied ways. As shown in Table~\ref{tab:res}, incorporating label definitions and requesting the top 6 labels significantly enhances the results for GPT-Vision, yet it adversely impacts LLaVA's performance. Thus, tailoring prompts to the specific characteristics and capabilities of each model may be necessary to achieve the best outcomes.
Furthermore, following~\cite{radford2021learning}, we adjusted the phrasing of the CLIP's input prompt from \textit{"The person in the red bounding box is [emotion label]"} to \textit{"A photo of a person in a red bounding who is [emotion label]."} Interestingly, this modification led to a decrease in the CLIP approach's F1 score to $13.76$.

\mypara{How does the number of people in the image impact the emotion recognition outcome?} 
We evaluate different methods based on the number of people in the image: one, two, or multiple people. As shown in Table \ref{table:num_people}, the precision, recall, and F1 score tend to decrease as the number of people increases. This reduction in performance can be attributed to the more complex situations that arise when there are more people in a scene~\cite{veltmeijer2021automatic}. Furthermore, NarraCap does not account for human interactions, and vision language models (VLMs) struggle with identifying the specific individual referred to in a prompt. This challenge is partly due to the models' limitations in interpreting visual markers (bounding boxes), which are crucial for distinguishing among multiple subjects in an image~\cite{wan2024contrastive}. 
For the EMOTIC model, which was trained on the training set, it was observed that while it surpassed the performance of NarraCap, a zero-shot approach, for images featuring a single person, it was less effective in handling images with two or more people. This discrepancy suggests that NarraCap demonstrates superior performance in more complex scenarios involving multiple individuals. 

\input{latex/tables/NumberOfPeople}

\mypara{How do age, gender, activity, environment, and physical signals affect the results?} 
One of the advantages of the NarraCap approach is that it provides a way to explicitly select image details to include for inference and perform ablations using the text representation. 
To assess the effect of gender, instead of using specific gender labels such as "a baby boy," "a baby girl,", etc. 
we only utilized the labels "a female" and "a male." Furthermore, to examine gender, we modified the label list to "a baby," "a kid," "an adult," and "an elderly person."
To investigate the impact of activity, environment, and physical signals, we excluded those components from the captions. The findings from each study, conducted on a set of 1000 images randomly selected from the validation set, are summarized in Table \ref{table:ablation}. This table reveals that the \textit{action} depicted in an image had the most significant impact on the outcome, followed by the \textit{environment}. These insights suggests that future research on image caption generation may focus on understanding image actions and contexts, as they are keys to create accurate and relevant captions.

%% file: latex/tables/Table1_pdf.tex
\begin{figure*}[t]
\centering
\includegraphics[width=1\linewidth]{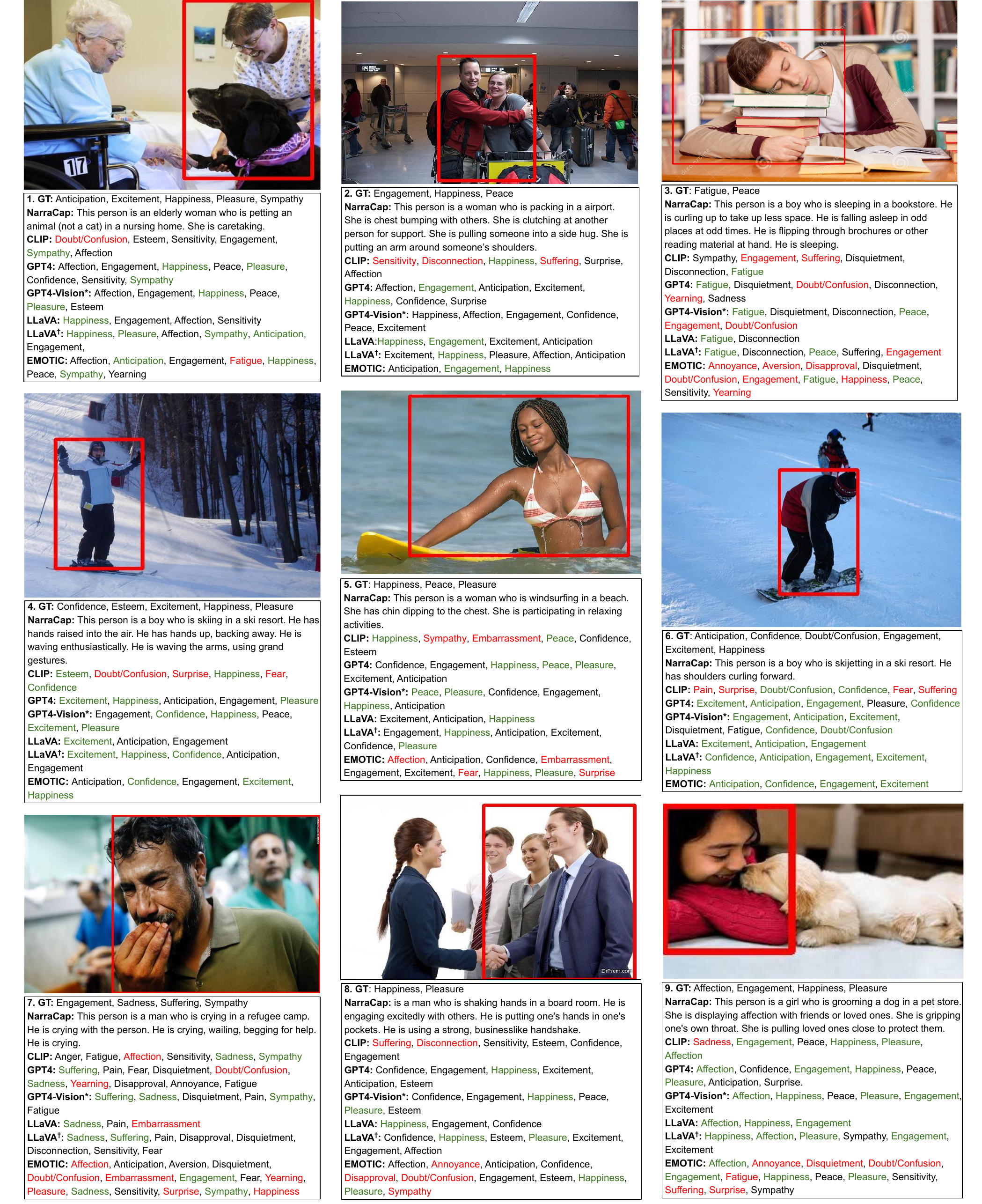}
\vspace{-7mm}
\caption{Qualitative analysis of EMOTIC images showcasing ground truth (GT) labels, NarraCap-generated captions, and inferred labels using (a) zero-shot methods: CLIP, GPT4 (NarraCap+GPT4), GPT4-Vision, LLaVA; and (b) trained methods: LLaVA finetuned on the validation set with augmentation (LLaVA\dag), and EMOTIC.}
  \label{tab:qualitative}
\vspace{-4mm}
\end{figure*}

%% file: latex/tables/ap_per_category.tex
\begin{table*}[ht]
\small
\centering
\resizebox{\textwidth}{!}{
\begin{tabular}{@{}l|lccccc@{}
  l 
  S[table-format=2.2(3)] 
  S[table-format=2.2(3)] 
  S[table-format=2.2(3)] 
  S[table-format=2.2(3)] 
 S[table-format=2.2(3)] 
 S[table-format=2.2(3)]
  @{}
  }

\toprule
& {Model} & {$\uparrow$Precision (\%)} & {$\uparrow$Recall (\%)} & {$\uparrow$F1 Score (\%)} & {$\downarrow$Hamming (\%)} & {$\uparrow$S-acc (\%)} \\
\midrule
\multirow{11}{*}{\rotatebox[origin=c]{90}{Zero shot}}
& Majority & $\textbf{\textcolor{Red}{11.41}}^{\textcolor{Red}{\pm 0.05}}$ & $23.08^{\pm 0.00}$ & $15.01^{\pm 0.05}$ & $17.24^{\pm 0.09}$ &  $0.76^{\pm 0.09}$ \\

& Rand6             & $16.90^{\pm 0.13}$ & $23.12^{\pm 0.34}$ & $\textbf{\textcolor{Red}{14.90}}^{\textcolor{Red}{\pm  0.15}}$ & $32.29^{\pm 0.07}$ &  $\textbf{\textcolor{Red}{0.00}}^{\textcolor{Red}{\pm 0.00}}$ \\

& Rand6-weighted    & $17.02^{\pm 0.15}$ & $23.17^{\pm 0.20}$ & $19.45^{\pm 0.17}$ & $22.67^{\pm  0.08}$ &  $0.01^{\pm 0.01}$ \\

& ExpNet + GPT4           & $24.94^{\pm 0.58}$ & $23.57^{\pm 0.27}$ & $22.29^{\pm 0.30}$ & $17.27^{\pm 0.09}$ &  $1.79^{\pm 0.13}$ \\
& NarraCap + GPT4     & $25.50^{\pm 0.32}$ & $33.37^{\pm 0.42}$ & $26.67^{\pm 0.30}$ & $21.26^{\pm 0.11}$ & $0.82^{\pm 0.09}$ \\

& NarraCap + Mistral & $17.56^{\pm 0.09}$ & $\textbf{\textcolor{Blue}{64.53}}^{\textcolor{Blue}{\pm 0.59}}$ & $23.89^{0.14\pm}$ & $\textbf{\textcolor{Red}{52.70}}^{\textcolor{Red}{\pm 0.24}}$ & $0.01^{\pm 0.01}$ \\ 




& CLIP          & $21.77^{\pm 0.19}$ & $28.58^{\pm 0.35}$ & $16.97^{\pm 0.18}$ & $31.72^{\pm 0.07}$ &  $\textbf{\textcolor{Red}{0.00}}^{\textcolor{Red}{\pm 0.00}}$ \\

& LLaVA           & $33.78^{\pm 0.86}$ & $21.38^{\pm 0.30}$ & $22.86^{\pm 0.32}$ & $\textbf{\textcolor{Blue}{15.02}}^{\textcolor{Blue}{\pm 0.08}}$ &  $0.99^{\pm 0.10}$ \\

& LLaVA*             & $27.77^{\pm 0.63}$ & $\textbf{\textcolor{Red}{18.51}}^{\textcolor{Red}{\pm 0.25}}$ & $19.58^{\pm 0.27}$ & $16.04^{\pm 0.08}$ &  $0.78^{\pm 0.09}$ \\

& GPT-4 vision       & $29.07^{\pm 0.37}$ & $27.48^{\pm 0.37}$ & $26.12^{\pm 0.28}$ & $16.72^{\pm 0.12}$ &  $\textbf{\textcolor{Blue}{1.90}}^{\textcolor{Blue}{\pm 0.14}}$ \\

& GPT-4 vision*       & $\textbf{\textcolor{Blue}{37.48}}^{\textcolor{Blue}{\pm 0.79}}$ & $38.35^{\pm 0.36}$ & $\textbf{\textcolor{Blue}{34.47}}^{\pm 0.35}$ & $16.95^{\pm 0.08}$ &  $0.67^{\pm 0.08}$ \\

\midrule
\multirow{6}{*}{\rotatebox[origin=c]{90}{Trained}}
& EMOTIC            & $25.02^{\pm 0.28}$ & $35.07^{\pm 0.49}$ & $28.83^{\pm 0.33}$ & $19.35^{\pm 0.14}$ &  $2.73^{\pm 0.17}$ \\
& Mistral-F (val set augmented) & $\textbf{\textcolor{Red}{18.01}}^{\textcolor{Red}{\pm 0.09}}$& $\textbf{\textcolor{Blue}{78.40}}^{\textcolor{Blue}{\pm 0.45}}$& $26.41^{\pm 0.13}$& $\textbf{\textcolor{Red}{54.16}}^{\textcolor{Red}{\pm 0.19}}$& $\textbf{\textcolor{Red}{0.00}}^{\textcolor{Red}{\pm 0.00}}$ \\

& LLaVA-F (train set)  & $\textbf{\textcolor{Blue}{54.27}}^{\textcolor{Blue}{\pm 1.42}}$ & $\textbf{\textcolor{Red}{16.81}}^{\textcolor{Red}{\pm 0.30}}$ & $\textbf{\textcolor{Red}{22.73}}^{\textcolor{Red}{\pm 0.37}}$ & $\textbf{\textcolor{Blue}{13.17}}^{\textcolor{Blue}{\pm 0.07}}$ & $1.72^{\pm 0.13}$\\

& LLaVA-F (val set)  & $32.55^{\pm 0.55}$ & $42.95^{\pm 0.42}$ & $34.42^{\pm 0.34}$ & $17.14^{\pm 0.09}$ & $0.78^{\pm 0.09}$ \\

& LLaVA-F (val set augmented)  & $38.71^{\pm 0.55}$ & $39.52^{\pm 0.42}$ & $\textbf{\textcolor{Blue}{36.83}}^{\textcolor{Blue}{\pm 0.37}}$ & $14.13^{\pm 0.08}$ & $\textbf{\textcolor{Blue}{2.90}}^{\textcolor{Blue}{\pm 0.17}}$ \\ 

& LLaVA-F (val set 100 samples)  & $31.36^{\pm 0.37}$ & $40.41^{\pm 0.40}$ & $33.85^{\pm 0.33}$ & $17.30^{\pm 0.09}$ & $1.28^{\pm 0.12}$ \\

\bottomrule

\end{tabular}

}

\caption{Performance metrics of various models using macro average. ``S-acc'' represents subset accuracy, ``-F'' indicates that this is a fine-tuned version, and ``*'' indicates prompt engineering (definitions + request 6 labels, ave. in validation set)}
\vspace{-8mm}
\label{tab:res}
\end{table*}

%% file: latex/tables/clip-map.tex
\begin{figure}[t]
\centering
\includegraphics[width=\linewidth]{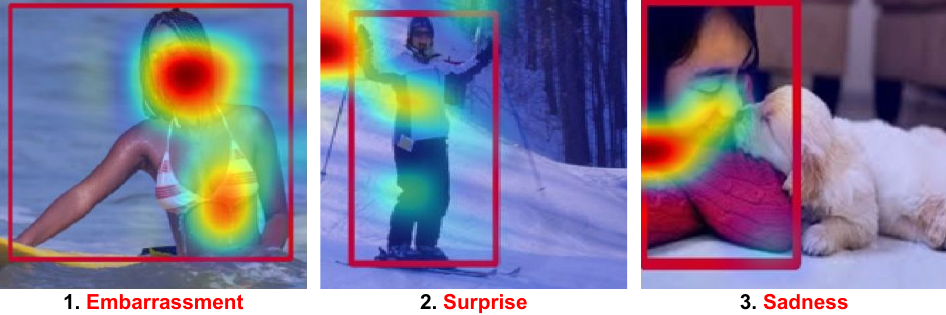}
\vspace{-7mm}
 \caption{CLIP saliency maps using Grad-Cam~\cite{selvaraju2017grad}. It identifies the regions in an input image that most significantly influence the classification score by leveraging the gradients of the score relative to the last convolutional layer's feature map.}
    \label{tab:clip-map}
\vspace{-4mm}
\end{figure}

%% file: latex/tables/human_set.tex
\begin{table}[t]
\centering
\resizebox{\linewidth}{!}{
\small
\begin{tabular}
{@{}cccccccc@{}}\toprule
& \multicolumn{4}{c}{Zero-shot} & & \multicolumn{2}{c}{Trained} 
 \\

\cmidrule{2-5} \cmidrule{7-8} 
& NC+GPT4 & HNC+GPT4 & GPT-Vis & LLaVA && LLaVA-f & EMOTIC  \\ 
\midrule
\textbf{F1}  & 26.19 & 34.17 & 35.79 & 27.08 && 42.14 & 26.50\\
 \bottomrule
\end{tabular}
}
\caption{Human-generated narrative captions + GPT-4 (HNC) vs. GPT-4 Vision (GPT-Vis) on negative emotion subset  \cite{yang2023contextual}}
\vspace{-8mm}
\label{tab:overview_f1_scores}
\end{table}

%% file: latex/tables/NumberOfPeople.tex

     

\begin{table}[t]

\small
    \centering
    \resizebox{\linewidth}{!}{
        \begin{tabular}{@{}l |ccccc|cc@{}}
        \toprule
         & & CLIP & NarraCap & GPT4-V & LLaVA & LLaVA-f & EMOTIC\\
         \midrule
        \multirow{3}{2em}{P}
        &1 & 22.07  & 26.44 & 37.53 & 32.06 & 41.27 & 27.72\\
        
        &2 & 21.52  & 26.06 & 36.64 & 36.10 & 38.07 & 23.36\\
        
        &>2 & 20.97 & 23.18 & 34.31 & 31.30 & 35.02 & 22.71\\
        \midrule
        \multirow{3}{2em}{R}
         &1 & 28.76  & 32.94 & 39.07 & 22.79 & 41.41 & 41.62\\
         
        &2 & 28.11 & 35.02 & 38.67 &  22.15 & 41.21 & 35.09\\
        &>2 & 27.45  & 31.74 & 36.54 & 18.39 & 35.66 & 28.51\\
        \midrule
        \multirow{3}{2em}{F1}
        &1 & 16.20 & 26.53 & 34.75 & 23.53 & 38.42 & 32.50\\
        
        &2 & 16.67  & 27.61 & 34.28 & 23.38 & 37.35 & 27.60\\
        &>2 & 16.81  & 25.01 & 32.57 & 19.77 & 33.34 & 24.60\\
        \bottomrule
        \end{tabular}
                            }
    \vspace{-2mm}
     \caption{Test set precision(p), recall(r), and F1 score when separated into 1, 2, and multiple people in the image. Here, NarraCap is paired with GPT-4}  
     
      \vspace{-8mm}            
\label{table:num_people}
\end{table}

%% file: latex/Sections/limitations.tex
\input{latex/tables/ablation}
The current study evaluating emotional theory of mind within the EMOTIC image dataset has its limitations. It primarily focused on various OpenAI models, including the zero-shot classifier CLIP, the large language model GPT-4, and the vision-language model GPT-4 Vision. Additionally, two open-source methods were examined: LLaVA (VLM) and Mistral (LLM). Although state-of-the-art zero-shot techniques were employed, fine-tuning some models (GPT-4, GPT-4 Vision) was not feasible due to their proprietary, closed-source nature. In addition, it is not possible to know the extent that the vision language models (i.e. GPT-4 Vision and LLaVA) were exposed to the EMOTIC test set. 

Another limitation is the absence of certain traditional emotion recognition models (e.g., emoticon) from our study, as their code was not made publicly available by the authors and our attempts to re-implement these models failed to replicate the reported results.

In addition, we noticed that the EMOTIC dataset, while being one of the most challenging image emotion datasets including context, also has small imperfections, including some bounding boxes that contain 2 people instead of only one (Fig.~\ref{tab:qualitative}.2). Although the case described is rare, a future study could evaluate on other datasets, e.g. one person emotion expression datasets without context, which is considered to be a simpler task.
For NarraCap, captions did not describe the social interactions or interactions with objects, which if added may increase performance. Moreover, for the activity and environment detection using CLIP, we performed a standard evaluation with a limited set of classes without a "don't know" or null class, resulting in some mis-captioning. 



%% file: latex/tables/ablation.tex
\begin{table}[t]
\small
    \centering
    \resizebox{\linewidth}{!}{

        \begin{tabular}{c c c c c| c c c}
        \toprule
        
        \multicolumn{5}{c}{\textbf{Ablation Settings}}
        \\

        age & gender & environment & action & physical signals & \textbf{F1} & \textbf{diff.} & \textbf{SE}\\
        \hline
        \checkmark & \checkmark & \checkmark & \checkmark & \checkmark & 29.67 & - & 0.78 \\
        
        - & \checkmark & \checkmark & \checkmark & \checkmark &28.41 & -1.26 & 0.73 \\

        \checkmark & - & \checkmark & \checkmark & \checkmark & 29.25 & -0.42 & 0.78 \\
        
        \checkmark & \checkmark & - & \checkmark & \checkmark &  27.27 & -2.4 & 0.78  \\

        \checkmark & \checkmark & \checkmark & - & \checkmark & 23.67 & -6 & 0.70 \\

        \checkmark & \checkmark & \checkmark & \checkmark & - &   29.47 & -0.2 & 0.85  \\

    \bottomrule
        \end{tabular}
    }
    \vspace{-2mm}
        \caption{Ablation study using NarraCap+GPT4 on 1000 randomly selected images from the validation set. Here, we present F1 scores along with their corresponding standard errors.
}
\label{table:ablation}
\vspace{-8mm}
\end{table}

%% file: latex/Sections/conclusion.tex
In this study, we delve into the potential of vision language models (VLMs) and large language models (LLMs) for assessing visual emotional theory of mind. Our findings reveal that the zero-shot approach of GPT-4 Vision with prompt engineering could outperform the trained EMOTIC model. Furthermore, our success in enhancing the performance beyond conventional methods by fine-tuning LLaVA, an open-source VLM, on a modest dataset underscores the profound potential of these models to comprehend human emotions.
Further research could explore improving the narrative captions by adding other contextual factors to the caption, such as human-object interactions and relationships with other people in the image. Further studies on the characteristics of emotionally comprehensive captioning, coupled with GPT or with human evaluators could be done. Additionally, enhancing the ability of VLMs to accurately recognize visual markers, such as bounding boxes, could significantly boost their performance.

%% file: root.bbl
\begin{thebibliography}{10}
\providecommand{\url}[1]{#1}
\csname url@rmstyle\endcsname
\providecommand{\newblock}{\relax}
\providecommand{\bibinfo}[2]{#2}
\providecommand\BIBentrySTDinterwordspacing{\spaceskip=0pt\relax}
\providecommand\BIBentryALTinterwordstretchfactor{4}
\providecommand\BIBentryALTinterwordspacing{\spaceskip=\fontdimen2\font plus
\BIBentryALTinterwordstretchfactor\fontdimen3\font minus \fontdimen4\font\relax}
\providecommand\BIBforeignlanguage[2]{{%
\expandafter\ifx\csname l@#1\endcsname\relax
\typeout{** WARNING: IEEEtran.bst: No hyphenation pattern has been}%
\typeout{** loaded for the language `#1'. Using the pattern for}%
\typeout{** the default language instead.}%
\else
\language=\csname l@#1\endcsname
\fi
#2}}

\bibitem{picard2000affective}
R.~W. Picard, \emph{Affective computing}.\hskip 1em plus 0.5em minus 0.4em\relax MIT press, 2000.

\bibitem{barrett2019emotional}
L.~Barrett~et al., ``Emotional expressions reconsidered: Challenges to inferring emotion from human facial movements,'' \emph{Psychological science in the public interest}, vol.~20, no.~1, pp. 1--68, 2019.

\bibitem{pantic2000expert}
M.~Pantic and L.~J. Rothkrantz, ``Expert system for automatic analysis of facial expressions,'' \emph{Image and Vision Computing}, vol.~18, no.~11, pp. 881--905, 2000.

\bibitem{schindler2008recognizing}
K.~Schindler~et al., ``Recognizing emotions expressed by body pose: A biologically inspired neural model,'' \emph{Neural networks}, vol.~21, no.~9, pp. 1238--1246, 2008.

\bibitem{barrett2011context}
L.~F. Barrett, B.~Mesquita, and M.~Gendron, ``Context in emotion perception,'' \emph{Current directions in psychological science}, vol.~20, no.~5, pp. 286--290, 2011.

\bibitem{barrett2017emotions}
L.~F. Barrett, \emph{How emotions are made: The secret life of the brain}.\hskip 1em plus 0.5em minus 0.4em\relax Pan Macmillan, 2017.

\bibitem{kosti2019context}
R.~Kosti~et al., ``Context based emotion recognition using emotic dataset,'' \emph{PAMI}, vol.~42, no.~11, pp. 2755--2766, 2019.

\bibitem{le2022global}
N.~Le~et al., ``Global-local attention for emotion recognition,'' \emph{Neural Computing and Applications}, vol.~34, no.~24, pp. 21\,625--21\,639, 2022.

\bibitem{mittal2020emoticon}
T.~Mittal~et al., ``Emoticon: Context-aware multimodal emotion recognition using frege's principle,'' in \emph{CVPR}, 2020, pp. 14\,234--14\,243.

\bibitem{ong2019computational}
D.~C. Ong~et al., ``Computational models of emotion inference in theory of mind: A review and roadmap,'' \emph{Topics in cognitive science}, vol.~11, no.~2, pp. 338--357, 2019.

\bibitem{lopez2016dependence}
D.~Lopez-Paz, ``From dependence to causation,'' \emph{arXiv}, 2016.

\bibitem{jang2022bc}
E.~Jang~et al., ``Bc-z: Zero-shot task generalization with robotic imitation learning,'' in \emph{CoRL}, 2022.

\bibitem{vaswani2017attention}
A.~Vaswani~et al., ``Attention is all you need,'' \emph{NeurIPS}, vol.~30, 2017.

\bibitem{brown2020language}
T.~Brown~et al., ``Language models are few-shot learners,'' \emph{Advances in neural information processing systems}, vol.~33, pp. 1877--1901, 2020.

\bibitem{chowdhery2022palm}
A.~Chowdhery~et al., ``Palm: Scaling language modeling with pathways,'' \emph{arXiv}, 2022.

\bibitem{antol2015vqa}
S.~Antol~et al., ``Vqa: Visual question answering,'' in \emph{ICCV}, 2015, pp. 2425--2433.

\bibitem{vinyals2015show}
O.~Vinyals~et al., ``Show and tell: A neural image caption generator,'' in \emph{CVPR}, 2015, pp. 3156--3164.

\bibitem{sap2019socialiqa}
M.~Sap~et al., ``Socialiqa: Commonsense reasoning about social interactions,'' \emph{arXiv}, 2019.

\bibitem{bisk2020piqa}
Y.~Bisk~et al., ``Piqa: Reasoning about physical commonsense in natural language,'' in \emph{AAAI}, vol.~34, no.~05, 2020, pp. 7432--7439.

\bibitem{li2022systematic}
X.~L. Li~et al., ``A systematic investigation of commonsense knowledge in large language models,'' in \emph{EMNLP}, 2022, pp. 11\,838--11\,855.

\bibitem{mao2022biases}
R.~Mao~et al., ``The biases of pre-trained language models: An empirical study on prompt-based sentiment analysis and emotion detection,'' \emph{IEEE Transactions on Affective Computing}, 2022.

\bibitem{sap2022neural}
M.~Sap~et al., ``Neural theory-of-mind? on the limits of social intelligence in large lms,'' \emph{arXiv}, 2022.

\bibitem{zhang2024vision}
J.~Zhang~et al., ``Vision-language models for vision tasks: A survey,'' \emph{PAMI}, 2024.

\bibitem{pratt2023does}
S.~Pratt~et al., ``What does a platypus look like? generating customized prompts for zero-shot image classification,'' in \emph{ICCV}, 2023, pp. 15\,691--15\,701.

\bibitem{long2023fine}
Y.~Long~et al., ``Fine-grained visual--text prompt-driven self-training for open-vocabulary object detection,'' \emph{IEEE Transactions on Neural Networks and Learning Systems}, 2023.

\bibitem{dhall2013finding}
A.~Dhall~et al., ``Finding happiest moments in a social context,'' in \emph{ACCV}.\hskip 1em plus 0.5em minus 0.4em\relax Springer, 2013, pp. 613--626.

\bibitem{huang2021emotion}
Y.~Huang~et al., ``Emotion recognition based on body and context fusion in the wild,'' in \emph{ICCV}, 2021, pp. 3609--3617.

\bibitem{lee2019context}
J.~Lee~et al., ``Context-aware emotion recognition networks,'' in \emph{ICCV}, 2019, pp. 10\,143--10\,152.

\bibitem{thuseethan2021boosting}
S.~Thuseethan~et al., ``Boosting emotion recognition in context using non-target subject information,'' in \emph{IJCNN}.\hskip 1em plus 0.5em minus 0.4em\relax IEEE, 2021, pp. 1--7.

\bibitem{mittel2023peri}
A.~Mittel and S.~Tripathi, ``Peri: Part aware emotion recognition in the wild,'' in \emph{ECCV 2022 Workshops}.\hskip 1em plus 0.5em minus 0.4em\relax Springer, 2023, pp. 76--92.

\bibitem{li2021human}
W.~Li~et al., ``Human emotion recognition with relational region-level analysis,'' \emph{IEEE Transactions on Affective Computing}, vol.~14, no.~1, pp. 650--663, 2021.

\bibitem{hoang2021context}
M.-H. Hoang~et al., ``Context-aware emotion recognition based on visual relationship detection,'' \emph{IEEE Access}, vol.~9, pp. 90\,465--90\,474, 2021.

\bibitem{yang2023context}
D.~Yang~et al., ``Context de-confounded emotion recognition,'' in \emph{CVPR}, 2023, pp. 19\,005--19\,015.

\bibitem{lindquist2015role}
L.~et~al., ``The role of language in emotion: Predictions from psychological constructionism,'' \emph{Frontiers in psychology}, vol.~6, p. 444, 2015.

\bibitem{lieberman2007putting}
M.~D. Lieberman~et al., ``Putting feelings into words,'' \emph{Psychological science}, vol.~18, no.~5, pp. 421--428, 2007.

\bibitem{lindquist2013s}
K.~A. Lindquist and M.~Gendron, ``What’s in a word? language constructs emotion perception,'' \emph{Emotion Review}, vol.~5, no.~1, pp. 66--71, 2013.

\bibitem{keen2015narrative}
S.~Keen and S.~Keen, ``Narrative emotions,'' \emph{Narrative Form: Revised and Expanded Second Edition}, pp. 152--161, 2015.

\bibitem{puglisi2019emotion}
B.~Puglisi and A.~Ackerman, \emph{The emotion thesaurus: A writer's guide to character expression}.\hskip 1em plus 0.5em minus 0.4em\relax JADD Publishing, 2019, vol.~1.

\bibitem{wang2019survey}
W.~Wang~et al., ``A survey of zero-shot learning: Settings, methods, and applications,'' \emph{TIST}, 2019.

\bibitem{liu2018generalized}
S.~Liu~et al., ``Generalized zero-shot learning with deep calibration network,'' \emph{NeurIPS}, 2018.

\bibitem{pourpanah2022review}
F.~Pourpanah~et al., ``A review of generalized zero-shot learning methods,'' \emph{PAMI}, 2022.

\bibitem{abderrahmane2018haptic}
Z.~Abderrahmane~et al., ``Haptic zero-shot learning: Recognition of objects never touched before,'' \emph{Rob. Auton. Syst.}, 2018.

\bibitem{socher2013zero}
R.~Socher~et al., ``Zero-shot learning through cross-modal transfer,'' \emph{NeurIPS}, 2013.

\bibitem{hu2022expansionnet}
J.~Hu~et al., ``Expansionnet v2: Block static expansion in fast end to end training for image captioning,'' \emph{arXiv}, 2022.

\bibitem{radford2021learning}
A.~Radford~et al., ``Learning transferable visual models from natural language supervision,'' in \emph{ICML}.\hskip 1em plus 0.5em minus 0.4em\relax PMLR, 2021, pp. 8748--8763.

\bibitem{smaira2020short}
L.~Smaira~et al., ``A short note on the kinetics-700-2020 human action dataset,'' \emph{arXiv}, 2020.

\bibitem{soomro2012ucf101}
K.~Soomro~et al., ``Ucf101: A dataset of 101 human actions classes from videos in the wild,'' \emph{arXiv}, 2012.

\bibitem{kuehne2011hmdb}
H.~Kuehne~et al., ``Hmdb: a large video database for human motion recognition,'' in \emph{ICCV}.\hskip 1em plus 0.5em minus 0.4em\relax IEEE, 2011, pp. 2556--2563.

\bibitem{puglisi2016urban}
B.~Puglisi and A.~Ackerman, \emph{The Urban Setting Thesaurus: A Writer's Guide to City Spaces}.\hskip 1em plus 0.5em minus 0.4em\relax JADD Publishing, 2016, vol.~5.

\bibitem{puglisi2016rural}
------, \emph{The Rural Setting Thesaurus: A Writer's Guide to Personal and Natural Places}.\hskip 1em plus 0.5em minus 0.4em\relax JADD Publishing, 2016, vol.~4.

\bibitem{de_Lima_Costa_2023_CVPR}
W.~de~Lima Costa~et al., ``High-level context representation for emotion recognition in images,'' in \emph{CVPR) Workshops}, June 2023, pp. 326--334.

\bibitem{jiang2023mistral}
A.~Q. Jiang~et al., ``Mistral 7b,'' \emph{arXiv}, 2023.

\bibitem{ainslie2023gqa}
J.~Ainslie~et al., ``Gqa: Training generalized multi-query transformer models from multi-head checkpoints,'' \emph{arXiv}, 2023.

\bibitem{child2019generating}
R.~Child~et al., ``Generating long sequences with sparse transformers,'' \emph{arXiv}, 2019.

\bibitem{beltagy2020longformer}
I.~Beltagy~et al., ``Longformer: The long-document transformer,'' \emph{arXiv}, 2020.

\bibitem{touvron2023llama}
H.~Touvron~et al., ``Llama: Open and efficient foundation language models,'' \emph{arXiv}, 2023.

\bibitem{min2021metaicl}
S.~Min~et al., ``Metaicl: Learning to learn in context,'' \emph{arXiv}, 2021.

\bibitem{ouyang2022training}
L.~Ouyang~et al., ``Training language models to follow instructions with human feedback, 2022,'' \emph{arXiv}, vol.~13, p.~1, 2022.

\bibitem{liu2022few}
H.~Liu~et al., ``Few-shot parameter-efficient fine-tuning is better and cheaper than in-context learning,'' \emph{NeurIPS}, vol.~35, pp. 1950--1965, 2022.

\bibitem{hu2021lora}
E.~J. Hu~et al., ``Lora: Low-rank adaptation of large language models,'' \emph{arXiv}, 2021.

\bibitem{dettmers2024qlora}
T.~Dettmers~et al., ``Qlora: Efficient finetuning of quantized llms,'' \emph{NeurIPS}, vol.~36, 2024.

\bibitem{liu2023visual}
H.~Liu~et al., ``Visual instruction tuning,'' \emph{arXiv}, 2023.

\bibitem{openai2023gpt4}
OpenAI, ``Gpt-4 technical report,'' 2023.

\bibitem{wei2022chain}
J.~Wei~et al., ``Chain-of-thought prompting elicits reasoning in large language models,'' \emph{NeurIPS}, vol.~35, pp. 24\,824--24\,837, 2022.

\bibitem{scikit-learn}
F.~Pedregosa~et al., ``Scikit-learn: Machine learning in {P}ython,'' \emph{Journal of Machine Learning Research}, vol.~12, pp. 2825--2830, 2011.

\bibitem{resnik1967context}
M.~D. Resnik, ``The context principle in frege's philosophy,'' \emph{Philosophy and Phenomenological Research}, vol.~27, no.~3, pp. 356--365, 1967.

\bibitem{selvaraju2017grad}
R.~R. Selvaraju~et al., ``Grad-cam: Visual explanations from deep networks via gradient-based localization,'' in \emph{ICCV}, 2017, pp. 618--626.

\bibitem{yang2023contextual}
V.~Yang~et al., ``Contextual emotion estimation from image captions,'' \emph{arXiv}, 2023.

\bibitem{veltmeijer2021automatic}
E.~A. Veltmeijer~et al., ``Automatic emotion recognition for groups: a review,'' \emph{IEEE Transactions on Affective Computing}, vol.~14, no.~1, pp. 89--107, 2021.

\bibitem{wan2024contrastive}
D.~Wan~et al., ``Contrastive region guidance: Improving grounding in vision-language models without training,'' \emph{arXiv}, 2024.

\end{thebibliography}
